# SMR-Net:Robot Snap Detection Based on Multi-Scale Features and Self-Attention Network

Kuanxu Hou

*Abstract*—In robot automated assembly, snap assembly precision and efficiency directly determine overall production quality. As a core prerequisite, snap detection and localization critically affect subsequent assembly success. Traditional visual methods suffer from poor robustness and large localization errors when handling complex scenarios (e.g., transparent or low-contrast snaps), failing to meet high-precision assembly demands. To address this, this paper designs a dedicated sensor and proposes SMR-Net, an self-attention-based multi-scale object detection algorithm, to synergistically enhance detection and localization performance. SMR-Net adopts an attention-enhanced multi-scale feature fusion architecture: raw sensor data is encoded via an attention-embedded feature extractor to strengthen key snap features and suppress noise; three multi-scale feature maps are processed in parallel with standard and dilated convolution for dimension unification while preserving resolution; an adaptive reweighting network dynamically assigns weights to fused features, generating fine representations integrating details and global semantics. Experimental results on Type A and Type B snap datasets show SMR-Net outperforms traditional Faster R-CNN significantly: Intersection over Union (IoU) improves by 6.52% and 5.8%, and mean Average Precision (mAP) increases by 2.8% and 1.5% respectively. This fully demonstrates the method's superiority in complex snap detection and localization tasks.

*Index Terms*—snap assembly,  snap detection and localization, object detection, multi-scale feature fusion, self-attention

## I. INTRODUCTION

Despite the widespread application of robotics in the manufacturing industry, the product assembly process still poses significant challenges to robotic systems, and this link is currently dominated by manual operations. This paper focuses on the key technical issues in robotic assembly, with particular emphasis on the assembly process of plastic parts with snap joints (hereinafter referred to as "snap assembly").

In snap assembly operations, the accurate recognition and localization of snaps serve as the core prerequisite for ensuring assembly success. Directly adopting unoptimized control strategies tends to result in overly aggressive robotic operations, which may lead to catastrophic slipping or structural damage of parts, thereby causing assembly failure or even system shutdown. This problem is particularly prominent in scenarios involving pose uncertainty and reliance on contact-based search operations [1][2].

Currently, visual localization methods based on standard cameras have demonstrated favorable performance in snap recognition and localization tasks under conventional scenarios, thanks to their ability to provide abundant appearance feature information. However, when snap parts are made of transparent materials or their surface color is highly similar to the background, traditional visual methods are constrained by imaging principles and struggle to effectively handle such complex working conditions. The dedicated sensor designed in this paper exhibits performance that depends solely on the surface texture depth of objects, unaffected by material transparency or background color similarity. Thus, it can effectively address the technical bottlenecks faced by traditional visual cameras in snap assembly localization. Based on the designed hardware system, this paper further proposes an object detection network integrating multi-scale features and a self-attention mechanism, aiming to enhance the accuracy and robustness of recognition and localization in snap assembly scenarios. In summary, the main contributions of this paper are as follows:

- We designed a novel sensor tailored for snap assembly scenarios, enabling high-precision recognition and localization of snaps in complex environments;
- We proposed an attention-based multi-scale network and integrated dilated convolution into its architecture to process the acquired multi-scale features prior to the Region Proposal Network (RPN) module;
- We developed an adaptive weight assignment network that can autonomously learn the importance of features at different scales and dynamically allocate weight coefficients, effectively enhancing the representational quality of fused multi-scale features and achieving state-of-the-art (SOTA) performance;
- We conducted comprehensive comparisons between the proposed algorithm and mainstream algorithms in terms of snap recognition, localization, and assembly, to evaluate the advanced performance of the proposed method;
- We performed ablation studies on each network component to fully demonstrate the compactness of the proposed algorithmic workflow.

## II. Related Work

### A. Tactile Sensor

The design and development of tactile sensors have formed a pattern of parallel development based on multiple principles, with mainstream technical routes including piezoelectric, resistive, electromagnetic, and optical approaches [3-6]. The core research objective of such sensors is to achieve accurate measurement of contact force and contact position within a specific area. However, traditional non-optical sensors generally face engineering application bottlenecks such as complex manufacturing processes and cumbersome wiring. In contrast, optical principle-based tactile sensors have been extensively studied due to significant advantages including simplified wiring, convenient manufacturing processes, and high spatial resolution. Schneiter and Sheridan [7] and Begej [8] utilized optical fibers to capture the light reflection characteristics of silicone surfaces and detected the variation patterns of optical fiber signals via cameras. Researchers such as Tar and Cserey [9] designed a sensor using a hollow hemispherical rubber cover as the contact medium, with a reflective surface integrated on the inner side of the cover's top. Three receivers are arranged at the bottom of the sensor, enabling the estimation of three-axis contact forces by measuring the reflected light signals from the deformed dome. Nevertheless, such sensors lack spatial measurement capability, a technical limitation that constitutes a critical constraint on the accuracy and reliability of snap positioning tasks.

### A. Object Detection Algorithm

Snap fasteners exhibit significant diversity in external morphology. Some snaps are not only miniature in size but also feature intricate texture structures. These characteristics make it challenging for deep learning-based object detection algorithms to fully extract effective features, thereby affecting detection accuracy and reliability. As one of the mainstream optimization approaches in the current object detection field, multi-scale feature fusion can effectively enhance the representation capability of fine textures and localization precision by integrating feature information from different network layers. For instance, FPN [10] constructs a top-down feature fusion pathway, significantly improving the semantic expression ability of low-level features; PANet [11] adds a bottom-up fusion branch on this basis, enabling bidirectional flow and in-depth interaction of feature information; YOLOF [12] designs efficient fusion units and expands the network receptive field, enhancing small object detection performance while ensuring computational efficiency.

Feature enhancement methods can improve the saliency of small objects and fine textures in feature maps. For example, dilated convolution [13] can capture contextual information while expanding the receptive field, strengthening the understanding of small objects; various attention mechanisms (SENet [14], DANet [15], CABM [16], GAM [17]) can guide the network to focus on regions of interest and reduce the interference of noise in the field of view.

However, small object detection still has limitations in adapting to extreme scale imbalance or irregular targets, and some enhancement strategies tend to introduce redundant information, leading to difficulties in training convergence. Therefore, achieving robust detection of small objects under the premise of ensuring efficiency remains an important practical research direction.

## III. Math

### A. Introduction to Sensor Hardware

*1) Principle of Gel Substrate:* The sensor designed in this paper aims to achieve high-precision measurement of the shape, appearance, and surface texture of target objects. The contact surface of the sensor with the target object is made of a transparent elastomer material, whose surface is coated with a silver powder coating with high reflectivity. When the target object is pressed against the elastomer surface, the elastic membrane undergoes adaptive deformation, thereby accurately replicating the three-dimensional topographical features of the object's surface. To capture this deformation information, a high-precision camera is arranged on the backside of the elastic membrane to record the object's surface contour presented on the elastic membrane. Based on the image data collected by the camera, a three-dimensional model of the target object's surface can be reconstructed through a three-dimensional reconstruction algorithm.

*2) Design of Elastic Substrates:* In this study, the elastic substrate of the sensor is designed with dimensions of 40mm×40mm and a thickness of 4mm. To effectively capture the subtle and complex texture information on the object surface while ensuring adaptability to surface textures of varying depths, the elastic substrate must possess excellent flexibility. The contact surface of the elastic substrate with the target object is coated with a fine silver powder coating. The diffuse reflection characteristic of this coating enables it to produce a uniform response to a wider range of surface normals, thereby providing favorable conditions for accurately capturing the overall topography of the object surface. Additionally, the minimum resolution of the sensor can reach 5 micrometers.

### B. Sensor Design and Fabrication

The main frame of the sensor is fabricated using 3D-printed nylon material. The sensor is primarily composed of a flange, a high-resolution industrial camera, a support frame, and a silver-plated elastic substrate. Among these components, the support frame is used to achieve the fixed installation of the camera and the elastic substrate. An acrylic guide plate is added between the elastic substrate and the support frame to ensure structural assembly accuracy. The elastic substrate conforms to the Neo-Hookean solid model, with a hardness close to Shore 00-45 and a shear modulus μ of 0.145 MPa. It is mounted at the central region of the bottom of the support frame. Under conditions of good illumination, when the elastic substrate deforms under compression, the light reflection direction of its surface silver-plated layer undergoes differential changes. This results in lower brightness in the areas corresponding to the object's edges, while the substrate surface in the contact region with the object exhibits higher brightness.

The distance between the center of the camera lens and the elastic substrate is set to 125 mm, with the elastic substrate having a thickness of 4 mm. In this study, a BFLY-U3-23S6C-C camera is adopted, which features a resolution of 1920×1200 pixels and is equipped with a Sony ICX249 1/1.2-inch progressive-scan CCD image sensor, enabling high-resolution acquisition of the contact surface's geometric topography. The camera is connected to a computer via a USB interface, facilitating efficient and rapid data transmission. After image data acquisition, the original image is cropped to retain only a central square region of 1000×1000 pixels.

*B. Algorithm Pipeline*

The network structure of Faster R-CNN [18] consists of a feature extraction module, a Region Proposal Network (RPN), and a target classification and bounding box regression module. Specifically, the feature extraction module is responsible for extracting semantic features and detailed features of the input image, while the RPN module generates target candidate regions based on the extracted features. To improve the accuracy of candidate regions, this study designs a convolutional feature fusion network with 5×5, 3×3, and 1×1 kernels to replace the 3×3 convolutional layer in the original RPN, enabling the fusion of three different-scale features output by the feature extraction module. Meanwhile, a self-attention mechanism module is embedded at an appropriate position in the convolutional network to enhance the feature extraction performance for Regions of Interest (ROIs). In summary, regarding the original Faster R-CNN, this study implements the aforementioned improvements to its feature extraction module and RPN module, respectively.

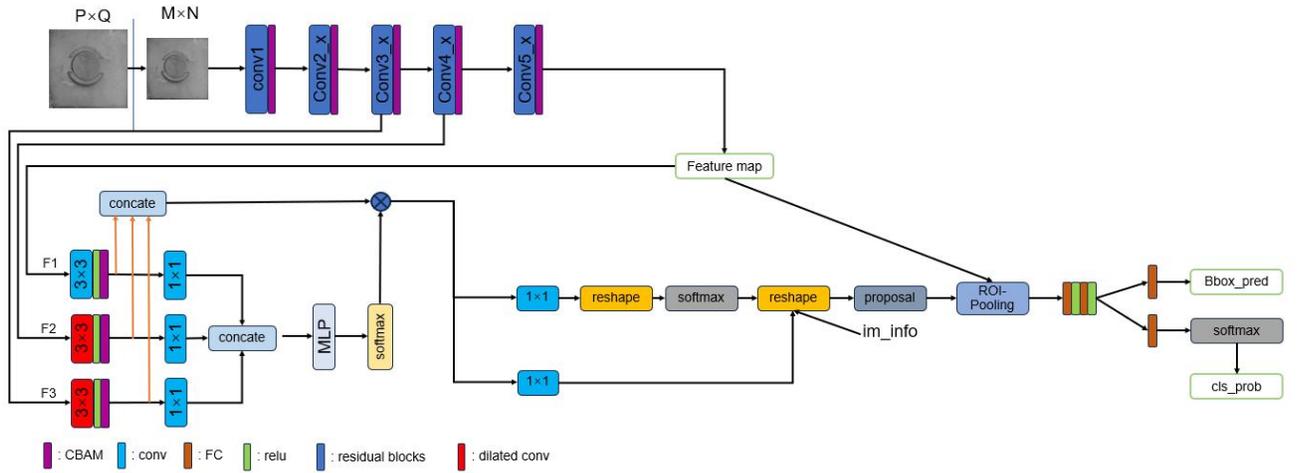

**Fig. 1.** Overview of the proposed SMR-Net pipeline. Based on Faster R-CNN, we integrate the Coordinate Attention Block Module (CABM) to enhance the target detection network's capability in detecting small objects and subtle textures. Specifically, conv1, conv2_x, conv3_x, conv4_x, and conv5_x in our backbone network are adapted from the convolutional layers and residual blocks of ResNet34 [19].

*A. Self-Attention Module*

Attention mechanisms can effectively enhance the feature representation capability of convolutional neural networks (CNNs). Therefore, we integrate the attention module—Coordinate Attention Block Module (CABM) [16]—after each residual block of the feature extraction network ResNet34. The specific implementation process of CABM is as follows:

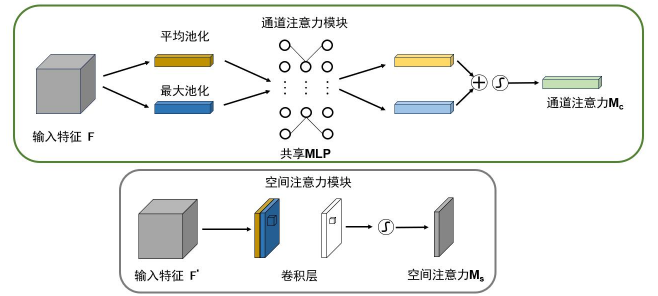

Fig. 2. Schematic diagram of the CABM [16] algorithm pipeline.

As shown in Fig. 1, the CBAM (Convolutional Block Attention Module) [16] consists of two sub-modules: CAM (Channel Attention Module) and SAM (Spatial Attention Module), which perform channel-wise and spatial-wise attention respectively. It can be seamlessly integrated into existing network frameworks. The Channel Attention Module preserves the channel dimension while compressing the spatial dimension, focusing on meaningful information in the input image. In contrast, the Spatial Attention Module maintains the spatial dimension and compresses the channel dimension, concentrating on the positional information of the target. These two modules (channel attention and spatial attention) can be combined in a serial manner, enabling more effective extraction of information from the image.

*B. Feature Extraction Network with self attention (SAFE-Net)*

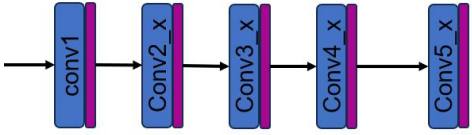

Fig. 3. Schematic diagram of the SAFE-Net feature extraction Network.

The feature extraction module serves as the foundation of the entire Faster R-CNN framework. The original network adopts VGG-16 [20] as its feature extraction backbone; however, this model suffers from an excessively large parameter scale, which restricts the further expansion of multi-scale network structures. To address this limitation, as illustrated in Fig. 3, this study employs ResNet-34 to replace VGG-16 for feature extraction tasks. Meanwhile, it outputs three feature maps of different scales for the RPN module, thereby enhancing the algorithm's detection accuracy for targets of varying sizes.

Specifically, the ResNet-34 network is structurally decomposed to obtain multiple convolutional layers and residual block units, including conv1, conv2_x, conv3_x, conv4_x, and conv5_x. An attention layer is appended after each convolutional layer and residual block. By adaptively focusing on key features and suppressing redundant information, the quality of extracted features is effectively improved. Ultimately, the three multi-scale feature maps output by this backbone network provide more discriminative feature support for the subsequent two-stage detection pipeline.

*C. Multi-Scale Feature Fusion Network (MSFF-Net)*

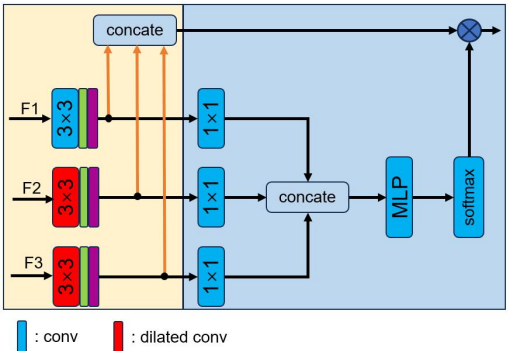

Fig. 4. Multi-scale feature fusion module.

Given the characteristics of some snaps in industrial scenarios, such as small sizes and subtle textures, conventional object detection algorithms fail to fully extract their features, resulting in low detection accuracy that is difficult to meet the requirements of practical applications. To address this issue, this study proposes a multi-scale feature fusion module, whose specific structure is illustrated in Fig. 4. As shown in the left part of Fig. 4, three feature maps of different scales (F1, F2, F3) output by the feature extraction network are first subjected to convolutional operations respectively: F1 is followed by a standard 3×3 convolution, while F2 and F3 are each followed by 3×3 dilated convolutions. Leveraging the property of dilated convolutions to expand the receptive field while preserving more detailed information, combined with the multi-scale feature fusion strategy, the localization regression accuracy of the RPN module can be effectively improved. Meanwhile, an attention module is integrated after each convolutional operation to further enhance the classification accuracy of object detection by strengthening the key features of the target and suppressing background interference.

The core reason why multi-scale feature fusion possesses unique advantages lies in the fact that feature maps at different levels contain differentiated information: detailed features of small objects are mainly embedded in low-level feature maps, while high-level feature maps are rich in semantic information. The original Faster R-CNN relies solely on high-level feature maps for object detection. After multiple downsampling operations on the input image, although the semantic features are abundant, the detailed features are severely lost. This leads to low confidence of candidate regions generated by the RPN, ultimately resulting in unsatisfactory small object detection performance.

*C. Re-weighting Network (RW-Net)*

There are significant differences in the contributions of features at different scales to the detection results. To utilize multi-scale information more efficiently, an adaptive weight learning module is designed to autonomously learn the optimal weights of features at each level. As shown in the right part of the network structure in Fig. 3., this module first performs channel-wise dimension compression and information integration on multi-scale features through 1×1 convolution. Subsequently, the processed features are input into a Multi-Layer Perceptron (MLP) for feature mapping and weight calculation. Finally, the weight coefficients of each feature layer are obtained through normalization via the softmax function. By weighted fusion of these weight coefficients with the corresponding scale feature maps, the final fused feature map that combines the advantages of features at all levels can be obtained, further enhancing the network's feature representation capability for targets of varying sizes.

IV. EXPERIMENTS

The experiments are conducted to evaluate the effectiveness of the proposed hardware and software system in the snap detection and installation process.

*A. Comparative Experiments with Mainstream Methods*

To evaluate the localization performance of the proposed method, the experiment adopts the Intersection over Union

(IoU) as the core metric, with the specific operational procedure as follows: Calculate the IoU between the optimal candidate bounding box output by the network and the manually annotated ground truth bounding box to quantify the model's localization accuracy. For the dataset, 1000 sensor images are collected for each of two different types of snaps to construct the experimental sample set; the final localization performance result is obtained by computing the mean value of IoU across all samples.

As a classic metric for measuring the overlap degree between two regions, IoU is widely used in computer vision tasks such as object detection and semantic segmentation. Its calculation method is as follows: By solving the ratio of the area of the overlapping region between the predicted candidate bounding box and the ground truth bounding box to the area of the union of the two regions, it intuitively reflects the consistency between the model's localization result and the real situation. The closer the ratio is to 1, the better the localization performance. Its calculation formula is given by:

$$IoU = \frac{P \cap Q}{P \cup Q} \quad (1)$$

TABLE I
IoU EXPERIMENTAL RESULTS OF DIFFERENT ALGORITHMS ON DIVERSE SNAP DATASETS

| Methods | Snaps | IoU (%) |
|---|---|---|
| Yolo V8[21] | A | 80.19 |
| | B | 82.66 |
| Fast R-CNN[22] | A | 80.58 |
| | B | 83.23 |
| Faster R-CNN[18] | A | 85.26 |
| | B | 86.32 |
| Ours | A | **91.78** |
| | B | **92.12** |

To verify the localization accuracy of the proposed algorithm, experiments were conducted for quantitative analysis using Intersection over Union (IoU) as the evaluation metric. For two types of snaps (Type A and Type B), 1000 sensor images were collected respectively to construct the experimental dataset. The IoU value was calculated between the predicted bounding box of the snap target and the ground truth bounding box in each image, and the final localization performance metric of each algorithm was obtained by summing and averaging the IoU results of all samples. As shown in the data in Table 1, the mean IoU of the proposed algorithm achieves the highest among the comparative methods. This result fully demonstrates that the proposed algorithm possesses superior localization accuracy and can more precisely realize the position prediction of snap targets.

TABLE II
COMPARISON OF mAP VALUES FOR SNAP RECOGNITION AMONG DIFFERENT ALGORITHMS

| Methods | Snaps | mAP (%) |
|---|---|---|
| Yolo V8[21] | A | 95.7 |
| | B | 95.1 |
| Fast R-CNN[22] | A | 95.5 |
| | B | 96.2 |
| Faster R-CNN[18] | A | 96.5 |
| | B | 97.9 |
| Ours | A | 99.3 |
| | B | 99.4 |

Due to differences in the installation methods of different types of snaps, accurate recognition of snap types is required. For the snap type recognition task, an experimental dataset was constructed by collecting 1000 images for each of Type A and Type B snaps. Mean Average Precision (mAP) was adopted as the evaluation metric: the detection precision of each image was calculated, and then summed and averaged to serve as the final type recognition performance result of each algorithm. As shown in the experimental data table, the proposed algorithm significantly outperforms other comparative algorithms in detection precision. Specifically, the recognition accuracy for Type A and Type B snaps is improved by 2.8% and 1.5% compared with Faster R-CNN, respectively. This fully verifies the effectiveness of the proposed improvement strategies in enhancing the ability to discriminate snap types.

To verify the application effectiveness of the proposed detection algorithm in actual snap installation scenarios, this study designed and conducted snap installation experiments. Two different types of snaps were selected, and 50 installation tests were performed for each type. The installation success rate was adopted as the core metric to evaluate the algorithm's performance. As shown in the experimental results in Table 1, the proposed algorithm achieves the optimal performance in terms of snap installation success rate. Relying on the proposed detection algorithm, the installation success rates of both Type A and Type B snaps reach 98%. This fully demonstrates that the algorithm can meet the requirements for accuracy and stability in practical industrial installation scenarios and possesses excellent engineering application value.

TABLE III
COMPARISON OF SNAP INSTALLATION SUCCESS RATES UNDER DIFFERENT ALGORITHM LOCALIZATION CONDITIONS

| Methods | Snaps | Success Rates |
|---|---|---|
| Yolo V8[21] | A | 88% |
| | B | 88% |
| Fast R-CNN[22] | A | 88% |
| | B | 88% |
| Faster R-CNN[18] | A | 90% |
| | B | 90% |
| Ours | A | **98%** |
| | B | **98%** |

*B. Ablation Study*

To systematically verify the effectiveness of the proposed algorithm, this study conducts ablation experiments on the core improved modules designed in the paper, namely the SA-FE feature extraction network, multi-scale feature fusion module, and reweighting network. The experiment adopts 1000 collected sensor images as the dataset. By constructing different algorithm structures (removing or retaining each core module respectively), independent training and testing are performed. The Intersection over Union (IoU) value is used as the performance evaluation metric to compare and analyze the contribution of each module to the detection accuracy, thereby verifying the effectiveness and necessity of each proposed algorithm structure.

TABLE IV
COMPARISON OF IoU VALUES AMONG DIFFERENT ALGORITHM STRUCTURES

| Ablation Studies | Snaps | IoU (%) |
|---|---|---|
| without SAFE-Net | A | 85.22 |
|  | B | 89.34 |
| without MSFF-Net | A | 86.32 |
|  | B | 88.37 |
| without RW-Net | A | 90.11 |
|  | B | 90.23 |
| Ours | A | **91.78** |
|  | B | **92.12** |

TABLE V
COMPARISON OF IoU VALUES AMONG DIFFERENT ALGORITHM STRUCTURES

| Ablation Studies | Snaps | mAP |
|---|---|---|
| without SAFE-Net | A | 97.2% |
|  | B | 97.1% |
| without MSFF-Net | A | 98.5% |
|  | B | 98.3% |
| without RW-Net | A | 98.9% |
|  | B | 99.1% |
| Ours | A | **99.3%** |
|  | B | **99.4%** |

*1) The Impact of the Feature Extraction Network (SAFE-Net):* The experimental results in Table IV and V demonstrate that after introducing the attention module into the network, both the recognition accuracy and localization precision of the model are significantly improved. This result verifies that the attention module combining spatial and channel dimensions can effectively focus on key valid information in images and suppress redundant background interference. Through adaptive enhancement and selection of features, it provides more discriminative feature support for subsequent target recognition and localization tasks, thereby improving the overall detection performance.

*2) The Impact of the Multi-Scale Feature Fusion Module (MSFF-Net):* The experimental results in Table IV and V show that regardless of whether the feature extraction network with an attention mechanism is adopted, the proposed multi-scale network exhibits superior detection performance compared to the single-scale network. This is because feature maps of different scales focus on information from different regions and levels in the image: low-level features are rich in detailed information, while high-level features contain more abundant semantic information. Relying solely on single-scale features (either middle-level or high-level) for detection tends to limit performance due to incomplete information representation. In contrast, multi-scale feature fusion can integrate the advantages of features at all levels, enabling comprehensive capture of target information and thus significantly improving detection performance. Additionally, the model performance is further enhanced after embedding the attention layer at an appropriate position in the network architecture. This indicates that the attention mechanism, by adaptively enhancing key target features and suppressing irrelevant background information, forms an effective performance complementarity with multi-scale fusion, thereby further strengthening the discriminative ability of the features.

*3) The Impact of the Reweighting Network (RW-Net):* In this experiment, a comparative model was constructed by removing the reweighting network and adopting the direct feature concatenation method. The experimental results show that under all test scenarios, the performance of the model with the introduced reweighting network is superior to that of the direct feature concatenation scheme. This result verifies that the proposed reweighting network can autonomously learn to assign optimal weights to each feature map, effectively excavate and fully utilize the important information of features at different scales. It avoids the problem of unbalanced feature information weights in simple concatenation, and ultimately enables the fused multi-scale features to possess more robust and discriminative representation capabilities, providing key support for improving detection performance.

## V. CONCLUSION

This paper analyzes the challenging scenarios faced by detection and localization tasks during the current robotic snap installation process. On this basis, a novel contact sensor hardware is designed, and an attention-based multi-scale network architecture (named SMR-Net) is proposed to improve the performance of snap detection and localization in challenging scenarios. The core design of SMR-Net consists of two aspects: first, a feature extraction network integrated with a self-attention mechanism to extract features at three different scales respectively; second, the integration of dilated convolution and a reweighting network to achieve feature fusion, whose effectiveness has been verified through ablation experiments. Compared with existing object detection algorithms, the proposed SMR-Net exhibits superior localization performance and snap recognition capability. Future research will focus on simplifying and accelerating the algorithm to realize its deployment and real-time operation on edge processors.